\documentclass[10pt,twocolumn,letterpaper]{article}

\usepackage[pagenumbers]{iccv} % To force page numbers, e.g. for an arXiv version

\definecolor{iccvblue}{rgb}{0.21,0.49,0.74}
\usepackage[pagebackref,breaklinks,colorlinks,allcolors=iccvblue]{hyperref}

\def\paperID{*****} % *** Enter the Paper ID here
\def\confName{ICCV}
\def\confYear{2025}

\title{Exploiting Information Redundancy in Attention Maps for Extreme Quantization of Vision Transformers}

\author{
Lucas Maisonnave$^*$\\
Université Paris‑Saclay CEA, List\\
F‑91120 Palaiseau, France\\
{\tt\small lucas.maisonnave@cea.fr}
\and
Karim Haroun$^*$\\
i3S / CNRS, Université Côte d’Azur\\
Sophia Antipolis, France\\
{\tt\small karim.haroun@etu.univ‑cotedazur.fr}
\and
Tom Pégeot\\
Université Paris‑Saclay CEA, List\\
F‑91120 Palaiseau, France\\
{\tt\small tom.pegeot@gmail.com}
}

\usepackage{multirow}
\usepackage{mathtools}
\begin{document}
\maketitle
\def\thefootnote{*}\footnotetext{These authors contributed equally to this work.}\def\thefootnote{\arabic{footnote}}
\begin{abstract}
Transformer models rely on Multi-Head Self-Attention (MHSA) mechanisms, where each attention head contributes to the final representation. However, their computational complexity and high memory demands due to MHSA hinders their deployment at the edge. In this work, we analyze and exploit information redundancy in attention maps to accelerate model inference. By quantifying the information captured by each attention head using Shannon entropy, our analysis reveals that attention heads with lower entropy, i.e., exhibiting more deterministic behavior, tend to contribute less information, motivating targeted compression strategies. Relying on these insights, we propose Entropy Attention Maps (EAM), a model that freezes the weights of low-entropy attention maps and quantizes these values to low precision to avoid redundant re-computation. Empirical validation on ImageNet-1k shows that EAM achieves similar or higher accuracy at $\leq$20\% sparsity in attention maps and competitive performance beyond this level for the DeiT and Swin Transformer models.
\end{abstract}

% % Le K.
% \begin{itemize}
%     \item Abstract : Reformuler pour que ça colle aux contributions (DONE)
%     \item Positionnement par rapport au SOTA concernant l'exploration de l'entropie pour les têtes d'attention (DONE)
%     \item Réécrire la conclusion (DONE)
    
%     \item Synchroniser les notations mathématiques et le modèle sur tout le papier (TODO)
%     \item Expériences : Terminer la partie expérimentale en analysant les résultats de la figure + le Tableau (TODO)
%     \item Introduction : Utiliser la figure 1 pour introduire l'hypothèse (TODO)
%     \item Refaire la figure 2 : Montrer comment l'entropie est calculée (cross samples) et justifier pourquoi (TODO)
% \end{itemize}

% % Le L.
% \begin{itemize}
%     \item Relire la partie quantification
%     \item Faire la partie divergence
%     \item section 4.4, rajouter une paragraphe 
% \end{itemize}

\section{Introduction}
\label{sec:intro}

Transformer models have achieved notable success in natural language processing (NLP) \cite{vaswani2017attention} and computer vision tasks \cite{dosovitskiy2020image, zhang2022segvit, strudel2021segmenter, carion2020end, Liu24SVit}, due to their ability to model long-range dependencies and handle variable-sized input sequences. Architectures such as BERT \cite{Devlin2019bert} and Vision Transformers (ViT) \cite{dosovitskiy2020image} have demonstrated state-of-the-art performance on a wide range of benchmarks. Their effectiveness is due to the Multi-Head Self-Attention (MHSA) mechanism, which allows the model to capture diverse contextual relationships through multiple attention heads operating in parallel.

\begin{figure}[!t]
    \centering
    \begin{minipage}{0.177\textwidth}
        \centering
        \includegraphics[width=\linewidth]{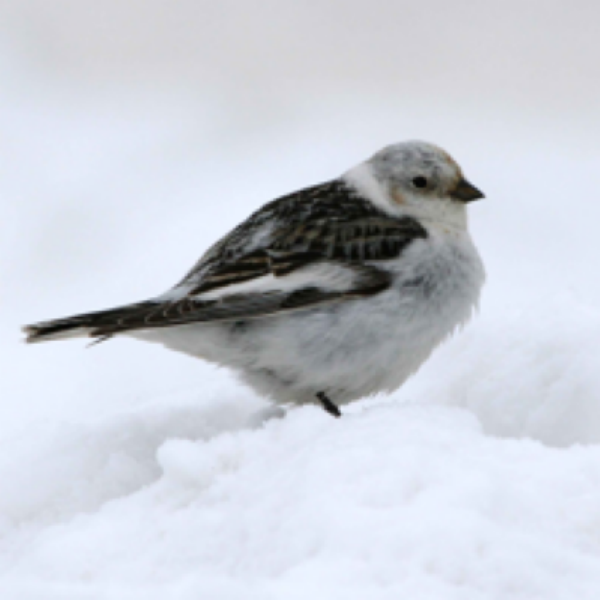}
        \subcaption{Original input image}
        \label{fig:a}
    \end{minipage}
    \hfill
    \begin{minipage}{0.23\textwidth}
        \centering
        \includegraphics[width=\linewidth]{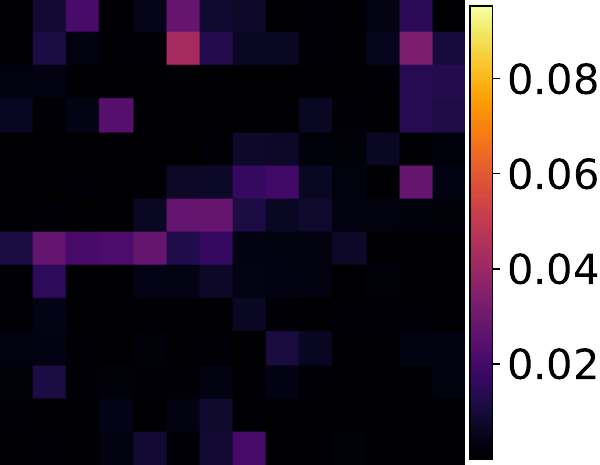}
        \subcaption{Attention head 1}
        \label{fig:b}
    \end{minipage}
    
    \begin{minipage}{0.23\textwidth}
        \centering
        \includegraphics[width=\linewidth]{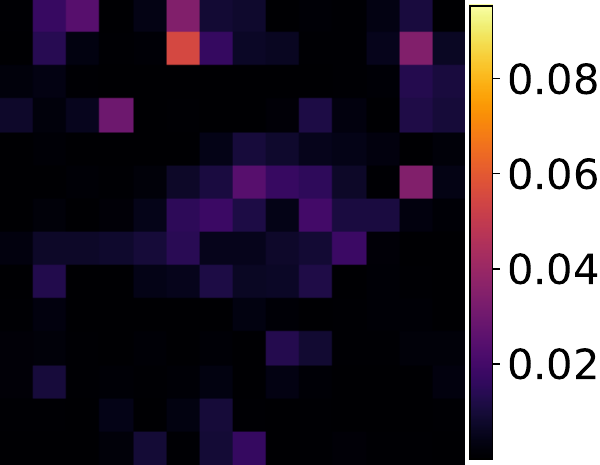}
        \subcaption{Attention head 2}
        \label{fig:c}
    \end{minipage}
    \hfill
    \begin{minipage}{0.23\textwidth}
        \centering
        \includegraphics[width=\linewidth]{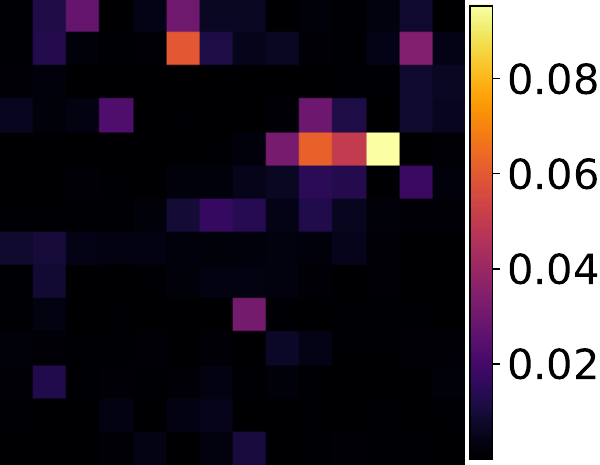}
        \subcaption{Attention head 3}
        \label{fig:c}
    \end{minipage}

    \caption{Visualization of the input image and the corresponding attention maps from the last layer’s attention heads of DeiT-Tiny.}
    \label{fig:visualizations}
\end{figure}

However, Transformers impose significant computational and memory demands, essentially due to MHSA \cite{tang2024survey}. Computing attention requires pairwise interaction of all token embeddings, resulting in a quadratic complexity $O(N^2)$ with respect to the sequence length, where $N$ is the number of tokens. This hinders deployment in environments with limited resources or tight latency requirements.

To mitigate these limitations, recent research has attempted to reduce complexity by designing more efficient models. Amongst the developed techniques, two main categories of complexity reduction stand out. The first involves token‑level sparsity \cite{haurum2023whichtokens}, in which attention is selectively applied only to a subset of tokens. The second is quantization \cite{liu2021post, li2022q, yuan2022ptq4vit, li2022patch, xiao2023patch, li2023repq}, where model weights and activations are encoded with lower numerical precision (e.g., from FP32 to 4‑bit integer), reducing both memory footprint and computational complexity.

% \begin{figure}[!t]
%     \centering
%     \begin{minipage}{0.177\textwidth}
%         \centering
%         \includegraphics[width=\linewidth]{images/input_image.pdf}
%         \subcaption{Original input image}
%         \label{fig:a}
%     \end{minipage}
%     \hfill
%     \begin{minipage}{0.23\textwidth}
%         \centering
%         \includegraphics[width=\linewidth]{images/attention_head_0.pdf}
%         \subcaption{Attention head 1}
%         \label{fig:b}
%     \end{minipage}
    
%     \begin{minipage}{0.23\textwidth}
%         \centering
%         \includegraphics[width=\linewidth]{images/attention_head_1.pdf}
%         \subcaption{Attention head 2}
%         \label{fig:c}
%     \end{minipage}
%     \hfill
%     \begin{minipage}{0.23\textwidth}
%         \centering
%         \includegraphics[width=\linewidth]{images/attention_head_2.pdf}
%         \subcaption{Attention head 3}
%         \label{fig:c}
%     \end{minipage}

%     \caption{Visualization of the input image and the corresponding attention maps from the last layer’s attention heads of DeiT-Tiny.}
%     \label{fig:visualizations}
% \end{figure}

\begin{figure*}[!t]
    
    \centering
    \includegraphics[width=\textwidth]{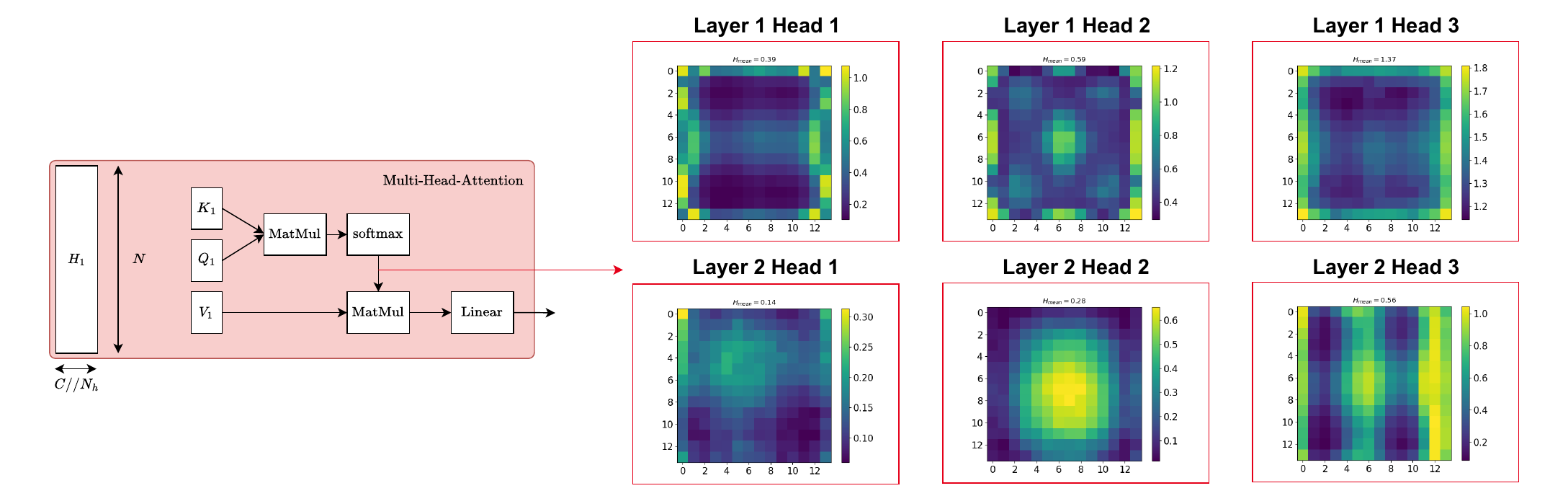}
    \caption{Entropy Attention Maps (EAM) of the $CLS$ token of a DeiT-Tiny computed on 5\% of ImageNet-1K. With this approach we can visualize differences between heads and the way they actually use information through attention.}
    \label{fig:viz_eam}
\end{figure*}

In this work, we focus on precision reduction of attention heads within the MHSA module, relying on two main observations. First, our visual analysis of individual attention maps, shown in Figure \ref{fig:visualizations} reveals that their weights frequently focus on small, localized regions of the input space rather than uniformly distributed across all positions. This spatial concentration suggests that a significant part of the attention computations may be redundant, as many attention weights contribute minimally to the context of the image. Second, we hypothesize that this redundancy can be quantified via Shannon entropy, i.e., heads exhibiting lower entropy, indicating limited variation across inputs, may be frozen and quantized to extremely low bit widths (as low as 4 bits) without affecting model performance, since their weights remain stable during inference. To test this hypothesis, we estimate for each head in every layer the entropy over the training dataset of the weights of the attention map. The resulting entropy values, as shown in Figure \ref{fig:viz_eam}, allow us to identify heads with consistently low variability.

%discussion sur l'observation et petite analyse

Relying on this observation, we develop a compression strategy that partially fixes (freezes) the attention weights of low-entropy attention maps and applies low-precision quantization. Despite removing their dynamic computation during inference, we retain the representational diversity of the high-entropy attention heads. Our contributions are summarized as follows:

\begin{itemize}
    \item We apply an entropy-based measure to quantify the information of each attention head, revealing that all attention heads exhibit a variable entropy across inputs.
    \item We propose a model that partially fixes the attention weights of low-entropy attention maps during inference and applies 4-bit quantization, reducing the computational complexity and memory demands without altering the model performance.
    \item We conduct extensive experiments on ImageNet-1K across various ViT architectures against state-of-the-art, and validate these results with an ablation study.
    % \item We extend our approach to Vision-Language Models (VLMs) where context windows are larger, illustrating its approach beyond image classification. 
\end{itemize}

% Our method extends existing quantization strategies by introducing a mechanism that selectively freezes parts of the attention mechanism based on their measured entropy. Specifically, we identify attention heads exhibiting low entropy, indicating limited variability across inputs, and freeze their attention maps during inference. All attention computations are then uniformly quantized to 4-bit precision, including both frozen and unfrozen heads.

% \begin{itemize}
%     \item Visualiser les cartes entropiques des autres tokens (Le L.)
%     \item Regarder le SOTA pour le mask (Le K., Le L.) 
%     \item Tester la méthode actuelle sur les couches spécifiques (Le K.) --quant_model.py
%     \item draw_entropy_map() pour plot les têtes d'attention
% \end{itemize}

\section{Related Works}
\label{sec:Related works}
% ViT/DeiT
% Entropy
% Quantification
% Fixed attention

\subsection{Vision Transformers}

As discussed in the Introduction, ViTs introduced by Dosovitskiy et al. \cite{dosovitskiy2020image} rely on the self-attention mechanism that computes contextual relationships between all tokens, where each token's representation is generated through learned query, key, and value projections followed by softmax-weighted aggregation across the entire token sequence. Building upon ViT, subsequent work addressed efficiency and scalability limitations. DeiT \cite{touvron2021deit} introduced distillation strategies to reduce training resource requirements, while Swin Transformer \cite{liu2021swin} proposed hierarchical feature maps and local-window attention to lower computational complexity. Despite these optimizations, the quadratic complexity of self-attention relative to the number of tokens in the sequence, coupled with high parameter counts, sustained significant computational and memory demands. 

To mitigate these constraints, researchers have designed efficient architectures that target low computation and memory during inference. Among these, Swin Transformer \cite{liu2021swin} introduced a hierarchical design using shifted window partitioning to efficiently limit attention computation to local regions. Pyramid Vision Transformer (PVT) \cite{wang2021pyramid} adopted a progressive shrinking pyramid structure to handle high-resolution inputs with reduced computation.

Besides, specific model compression strategies have been developed for integration into existing architectures. These are broadly categorized into two families: token reduction, which exploits sparsity, and quantization, which reduces numerical precision. Token reduction methods decrease the sequence length processed by the ViT, either by pruning redundant tokens \cite{liang2022evit, Tang23DToP, proust2025step} or merging semantically similar tokens \cite{bolya2023token, kim2024tokenfusion, haroun2024leveraging, haroun2025dynamic}. These approaches reduce computational complexity, but leave model parameters uncompressed, resulting in comparable memory demands for weights. In this paper, we focus on quantization techniques that reduce the precision of both weights and activations. These will be detailed in the following section.

\subsection{Quantization}

Quantization reduces the numerical precision of weights and activations in neural networks, typically from 32-bit floating-point to lower bit-width fixed-point or integer representations \cite{gholami2022survey}. Early quantization methods were developed for CNNs \cite{he2016deep, howard2019searching}, these methods include DoReFa-Net \cite{zhou2016dorefa} which approximates gradients in quantization-aware training by straight-through estimator (STE) \cite{bengio2013estimating}, and PACT \cite{choi2018pact} which propose parameterized clipping for activation quantization. Other works adopted non-uniform quantization \cite{Li2020Additive, liu2022nonuniform, yvinec2023powerquant}, and mixed-precision quantization \cite{micikevicius2018mixed, wang2019haq, xiao2023patch, ranjan2025mix}, where different bit widths are assigned to weights and activations based on their sensitivity, typically determined through pre-computed measures prior to inference.

Recent research has extended quantization methods to Vision Transformers. Ranking Loss \cite{liu2021post} preserves the relative order of quantized attention maps through a dedicated loss function. Q-ViT \cite{li_q-vit_2022} implements differentiable quantization, treating bit widths and scaling factors as learnable parameters during optimization. PTQ4ViT \cite{yuan2022ptq4vit} introduces twin uniform quantization coupled with a Hessian-guided metric for scaling factor selection. To handle non-linear operations, FQ-ViT \cite{lin2021fq} employs powers-of-two scaling for LayerNorm and logarithmic integer quantization for Softmax outputs. RepQ-ViT \cite{li2023repq} decouples quantization from inference pipelines to manage extreme activation distributions in LayerNorm and Softmax layers. Finally, PSAQ-ViT \cite{li2022patch} enables data-free quantization by leveraging patch similarity metrics.

%%%%MANQUE POSITIONNEMENT%%%%%%%

\section{Motivations}
\label{sec:Motivations}
% Ici on cherche à montrer l'intérêt d'étudier les têtes d'attention indépendemment
% 

% \subsection{Transformers and complexity}

% %% Le K.
% \textbf{I. Transformers: Definition and complexity}

% \begin{itemize}
%     \item General definitions on transformer models
%     \item the need for complexity reduction
%     \item Transition to the hypothesis
% \end{itemize}

Consider a tokenized input sequence $\mathbf{X} \in \mathbb{R}^{N \times d_e}$, with $N$ tokens each of dimension $d_e$, obtained by partitioning an RGB image $I \in \mathbb{R}^{H \times W}$ into non-overlapping patches $\mathbf{I_p} \in \mathbb{R}^{H_p \times W_p}$. The Transformer architecture processes this through two modules, a Multi-Head Self-Attention (MHSA) and a Multi-Layer Perceptron (MLP). The MHSA mechanism first projects the input sequence $X$ into three distinct representations query (Q), key (K) and values (V) through linear projections:

\begin{equation}
    Q = \mathbf{X}W_Q,\quad K = \mathbf{X}W_K,\quad V = \mathbf{X}W_V
\end{equation}

where $Q, K, V \in \mathbb{R}^{N \times d_e}$, and $W_Q$, $W_K$, $W_V$ $\in \mathbb{R}^{d_e \times d_e}$ are learned projection matrices for queries, keys, and values, respectively. Attention is then computed using the scaled dot-product formulation:

\begin{equation}
    A = \text{Softmax}\left( \frac{QK^\top}{\sqrt{d_e}} \right) \in \mathbb{R}^{N \times N}
\end{equation}

This attention map is used to aggregate the value vectors:

\begin{equation}
    O = AV \in \mathbb{R}^{N \times d_e}
\end{equation}

A final linear projection is applied to the output:

\begin{equation}
    \hat{O} = OW^{\text{proj}}, \quad W^{\text{proj}} \in \mathbb{R}^{d_e \times d_e}
\end{equation}

Each token is then independently passed through an MLP with two fully connected layers:

\begin{equation}
    \text{MLP}(\mathbf{X}) = \sigma(\mathbf{X}W_1)W_2
\end{equation}

where $W_1 \in \mathbb{R}^{d_e \times 4d_e}$ and $W_2 \in \mathbb{R}^{4d_e \times d_e}$ are learned weight matrices and $\text{MLP}(\mathbf{X}) \in \mathbb{R}^{N \times d_e}$ is the output of the Transformer layer.

As for the computation complexity expressed in FLOPs, each module's complexity breaks down as follows:

\begin{align}
    \Phi_{\text{MHSA}}(N, d_e) &= 4Nd_e^2 + 2N^2 d_e \\
    \Phi_{\text{MLP}}(N, d_e) &= 8Nd_e^2
\end{align}

Finally, the total computational cost of a Transformer layer can be decomposed as follows:

\begin{align}
    \Phi_{\text{Layer}}(N, d_e) &= \Phi_{\text{MHSA}}(N, d_e) + \Phi_{\text{MLP}}(N, d_e) \\
    &= 12Nd_e^2 + 2N^2 d_e
    \label{eq:total_flops}
\end{align}

The expression in Eq (\ref{eq:total_flops}) reveals that Transformers display quadratic complexity with respect to the sequence length $N$, as shown with the term $2N^2 d_e$. Besides, the MHSA mechanism requires the storage of the attention matrix $A \in \mathbb{R}^{N \times N}$, and the projection matrices ($W_Q$, $W_K$, $W_V$, $W^{\text{proj}}$) contribute by $O(d_e^2)$ parameters. Additionally, the MLP's weight matrices ($W_1$, $W_2$) introduce $O(d_e^2)$ parameters per layer. To alleviate this memory overhead, quantization emerges as a suitable optimization strategy.

\begin{figure*}[!t]
    
    \centering
    \begin{subfigure}[b]{0.23\textwidth}
        \centering
        \includegraphics[width=\textwidth]{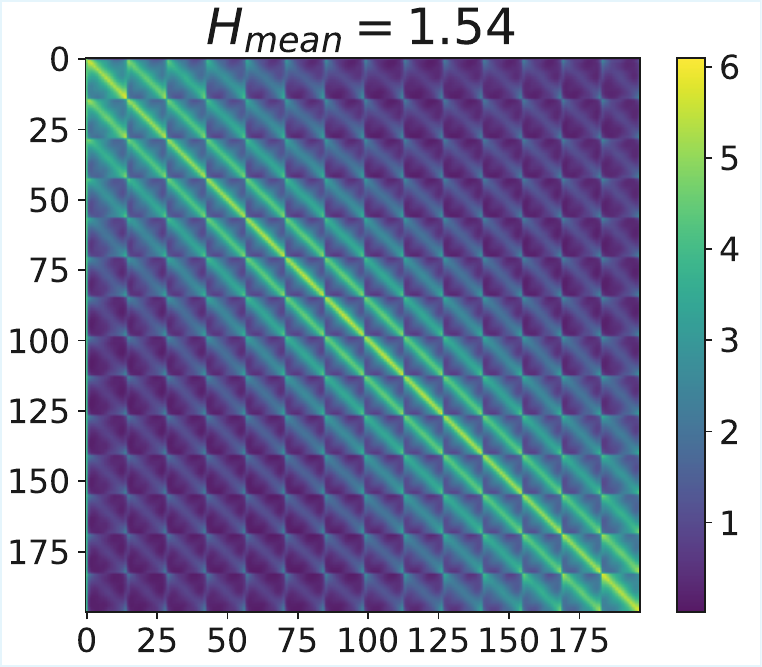}
        \caption{Layer 2 Head 2}
        \label{fig:image1}
    \end{subfigure}
    \hfill
    \begin{subfigure}[b]{0.23\textwidth}
        \centering
        \includegraphics[width=\textwidth]{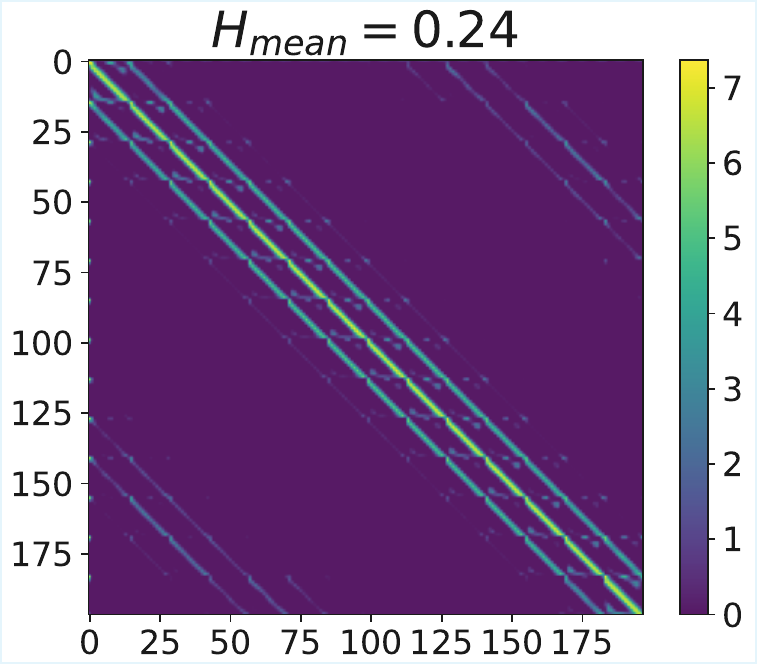}
        \caption{Layer 2 Head 3}
        \label{fig:image2}
    \end{subfigure}
    \hfill
    \begin{subfigure}[b]{0.23\textwidth}
        \centering
        \includegraphics[width=\textwidth]{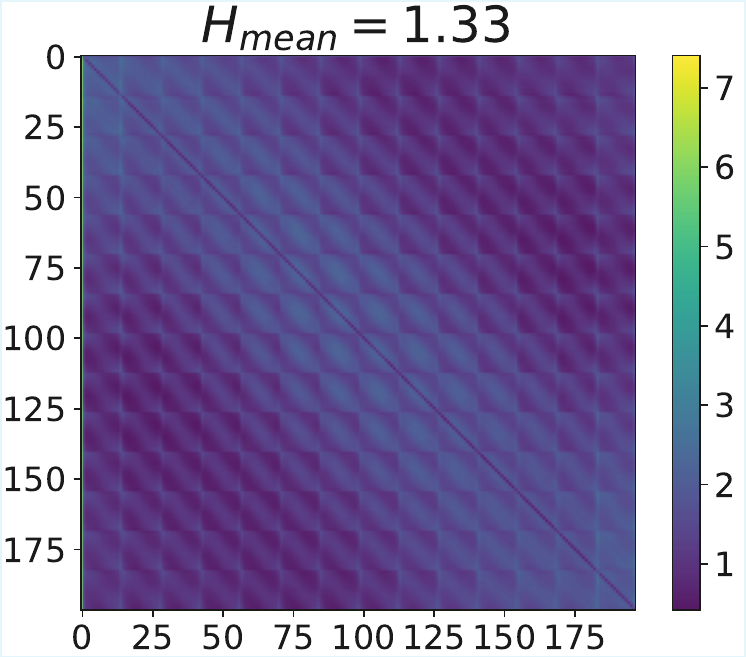}
        \caption{Layer 5 Head 1}
        \label{fig:image1}
    \end{subfigure}
    \hfill
    \begin{subfigure}[b]{0.23\textwidth}
        \centering
        \includegraphics[width=\textwidth]{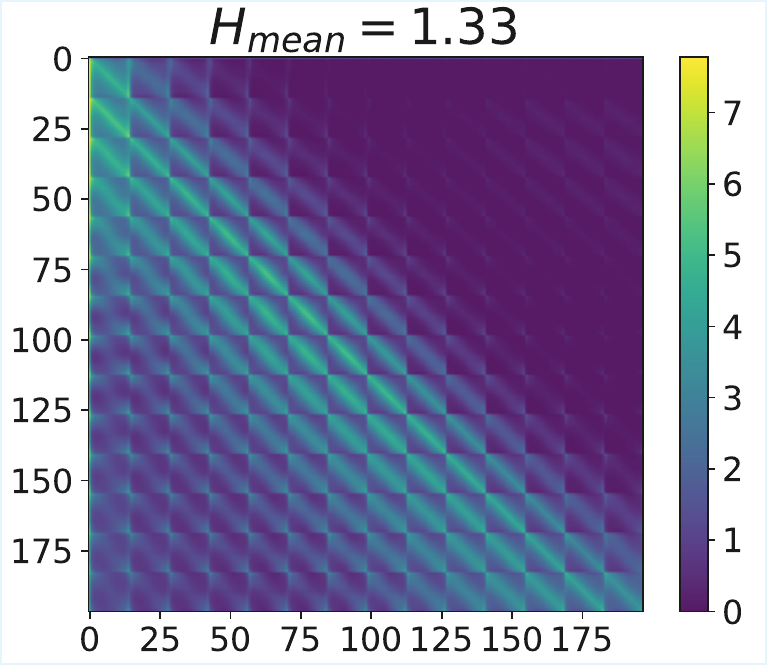}
        \caption{Layer 6 Head 3}
        \label{fig:image2}
    \end{subfigure}
    
    \caption{Full entropy maps of a DeiT-Tiny computed on 5\% of ImageNet-1K.}
    \label{fig:viz_full_eam}
\end{figure*}

As discussed in Section \ref{sec:intro}, attention weights are mainly concentrated to specific regions rather than uniformly distributed. This concentration creates redundancy, as many weights contribute minimally to the context of the input. Our approach is motivated by two hypotheses: (1) The redundancy in attention computations can be quantified by analyzing the entropy of attention weights across multiple input samples. (2) Attention weights that exhibit low entropy, demonstrating stable and predictable patterns across different inputs, can be frozen and quantized to low precision.

Specifically, for each attention head, we measure the entropy of its weight distributions over a dataset, which serves as an indicator of the weight stability. The details of this approach are formalized in the following section.

% \begin{figure*}[!t]
    
%     \centering
%     \begin{subfigure}[b]{0.23\textwidth}
%         \centering
%         \includegraphics[width=\textwidth]{images/head_full4.pdf}
%         \caption{Layer 2 Head 2}
%         \label{fig:image1}
%     \end{subfigure}
%     \hfill
%     \begin{subfigure}[b]{0.23\textwidth}
%         \centering
%         \includegraphics[width=\textwidth]{images/head_full5.pdf}
%         \caption{Layer 2 Head 3}
%         \label{fig:image2}
%     \end{subfigure}
%     \hfill
%     \begin{subfigure}[b]{0.23\textwidth}
%         \centering
%         \includegraphics[width=\textwidth]{images/head_full16.pdf}
%         \caption{Layer 5 Head 1}
%         \label{fig:image1}
%     \end{subfigure}
%     \hfill
%     \begin{subfigure}[b]{0.23\textwidth}
%         \centering
%         \includegraphics[width=\textwidth]{images/head_full19.pdf}
%         \caption{Layer 6 Head 3}
%         \label{fig:image2}
%     \end{subfigure}
    
%     \caption{Full entropy maps of a DeiT-Tiny computed on 5\% of ImageNet-1K.}
%     \label{fig:viz_full_eam}
% \end{figure*}

% \textbf{II. What motivates our work for Transformers}

% \begin{itemize}
%     \item Observations (Attention heads have different maps)
%     \item Hypothesis 1: Measuring the entropy of heads lead us to quantify information
%     \item Hypothesis 2: Quantize more aggressively some heads than others
%     \item Hypothesis 3: Entropy is a measure of uncertainty, low entropy indicates little variance in the attention values. Therefore, these values can be frozen to avoid recalculating them.
%     \item Applications : ViTs, (LLMs?)
% \end{itemize}

\section{Methodology}
\label{sec:Methodology}
% Le L.
\subsection{Entropy: Information and uncertainty}

% Attention mechanisms in neural networks are designed to establish relationships between tokens, thereby determining the relevance of one token to another. Attention maps, which visualize these relationships, have been leveraged to quantify the information content within models. By analyzing attention scores, researchers have been able to identify and prune or quantize less important parts of the model \cite{xiao2023patch}, thereby enhancing efficiency without significantly compromising performance. This underscores the pivotal role of attention scores in highlighting the critical components within a model.

Entropy, a fundamental concept in information theory, provides a robust metric to identify sensitive and useful parameters within a model. It has been widely used in various applications, including model quantization and regularization \cite{pereyra2017regularizing, maisonnave2024applying}, to optimize performance and reduce complexity. Entropy is defined as the amount of information contained in a probability distribution, representing the minimum number of bits required to encode the distribution without loss of information. Mathematically, the entropy $\mathcal{H}(X)$ of a random variable $X$ with probability distribution $p(x)$ is given by:

\begin{align}
\mathcal{H}(X) = - \sum_{x \in \mathcal{X}}p(x)\log_2 p(x), \, \text{with} \, X \sim p(x)
\end{align}

Entropy can also be interpreted as a measure of uncertainty. In this context, information is inversely related to the predictability of an event. An event that is highly uncertain or surprising carries more information as it challenges our existing knowledge and expectations. This property makes entropy a valuable tool for assessing the informational content and sensitivity of the model parameters.

By combining attention scores with entropy, we can gain deeper insight into model dynamics and better discern important information from redundant information. In this way, we can better spot useful computation instead of useful parameters. As depicted above, the attention mechanism is a computationally expensive component and quadratically increases with the number of tokens $N$, the purpose of this paper is to find which computation in the attention mechanism is redundant and can be avoided using entropy.

\begin{figure*}[!t]
    
    \centering
    \begin{subfigure}[b]{0.30\textwidth}
        \centering
        \includegraphics[width=\textwidth]{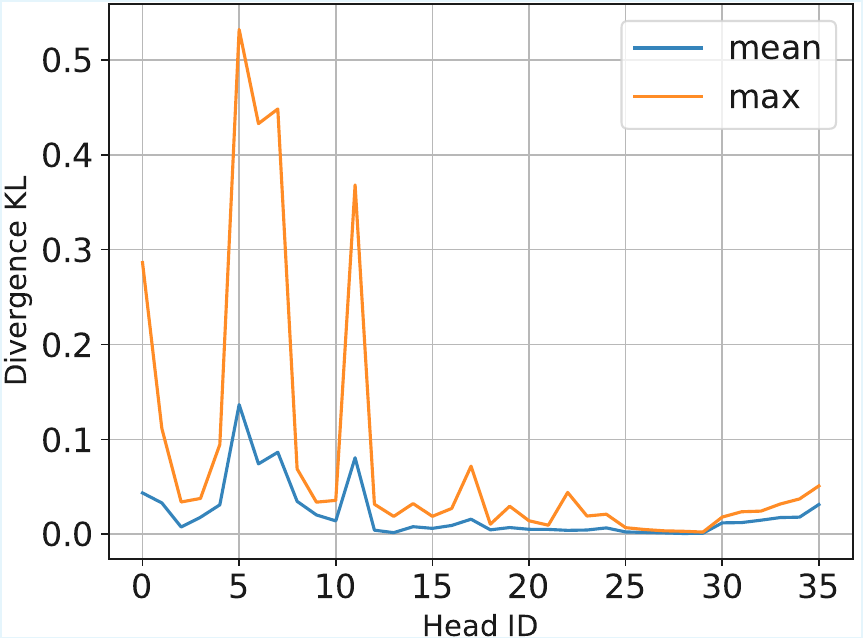}
        \caption{}
        \label{fig:image1}
    \end{subfigure}
    \hfill
    \begin{subfigure}[b]{0.33\textwidth}
        \centering
        \includegraphics[width=\textwidth]{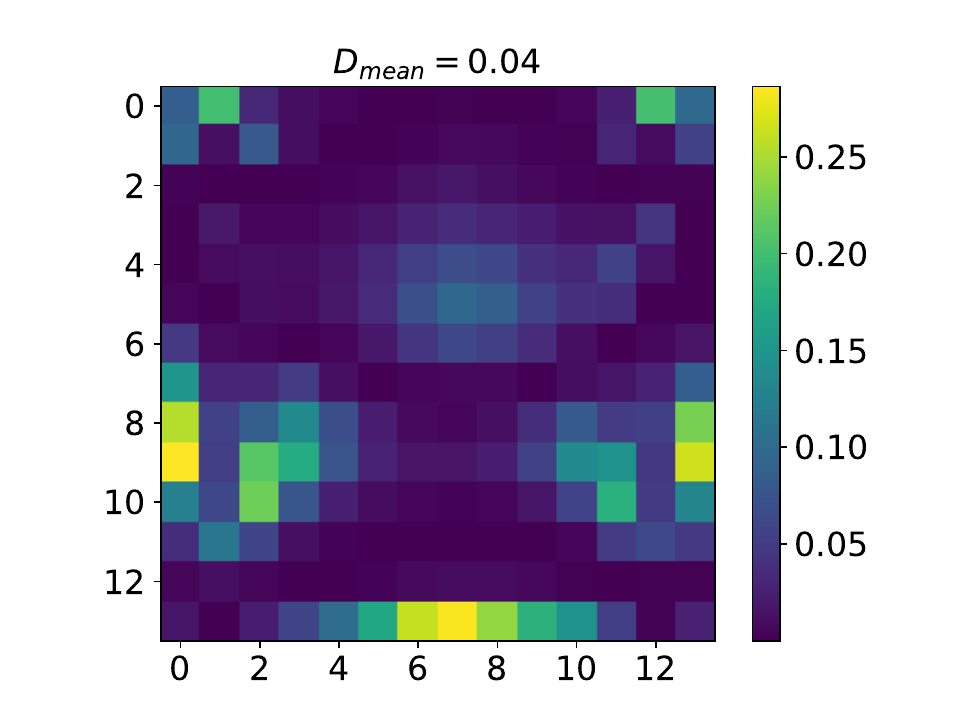}
        \caption{}
        \label{fig:image1}
    \end{subfigure}
    \hfill
    \begin{subfigure}[b]{0.33\textwidth}
        \centering
        \includegraphics[width=\textwidth]{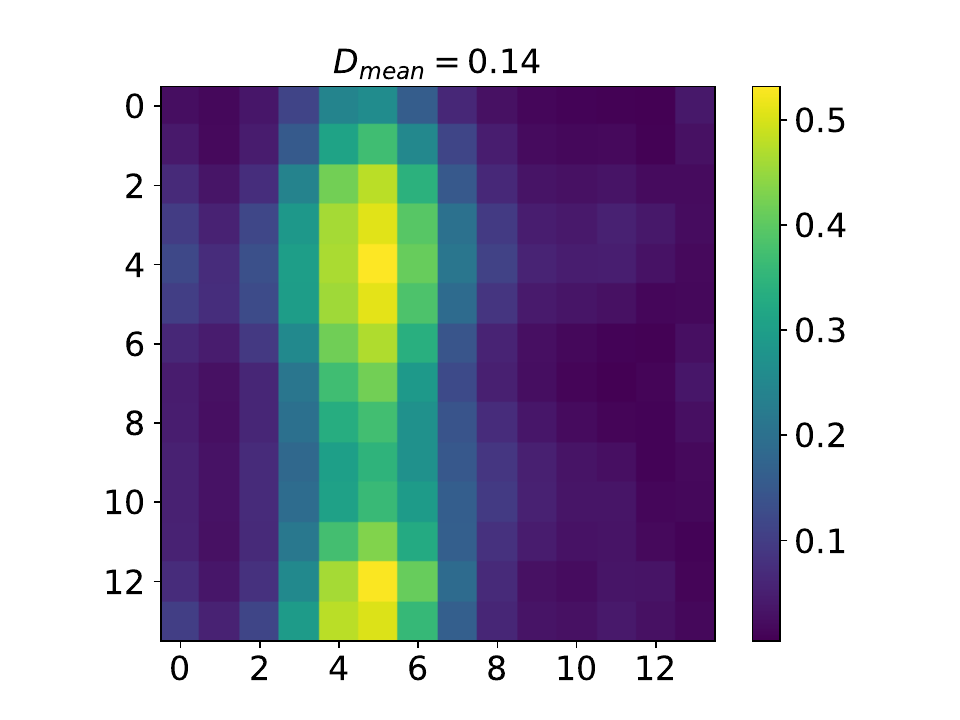}
        \caption{}
        \label{fig:image2}
    \end{subfigure}
    
    \caption{Visualization of $KL$ divergence of  attention weights distributions between the non-quantized model $p_{\text{fp32}}(x)$ and the 4-bit quantized model $p_q(x)$ on DeiT-Tiny, computed on 5\% of ImageNet-1K. (a) Evolution of divergence of the $CLS$ token over heads of the model. (b-c) Divergence maps on the $CLS$ token of a DeiT-Tiny for Layer 1 Head 1, and Layer 2 Head 3 respectively.}
    \label{fig:viz_div}
\end{figure*}

\subsection{Entropy Attention Maps}

\subsubsection{Definition}

To better quantify the behavior of attention maps and their uncertainty, we compute the entropy of each of their weights. We take every weight as a random variable with some distribution $p$ taking the values in $[0,1]$.

We define the attention map of the layer $l$ and the head $h$ of an image $m$ by:

\begin{align}
A^{m}_{l,h} = \text{softmax}\left (\frac{K^{m}_{l,h}{Q^{l,m}_{h}}^T}{\sqrt {d_l}} \right)
\end{align}

$A^{m}_{l,h} \in \mathbb{R}^{(N+1)\times (N+1)}$, since we add the Self-Attention of the $CLS$ token. We estimate the distribution $p$ of each attention weight $i$ of this random variable with a histogram quantized in $b=8$ bits. This way, we decompose the distribution into 256 values between 0 and 1 due to the softmax function.

\begin{align}
p^{i}_{l, h}(k) &= \frac{1}{M}\sum_m^M \left(A^{m}_{l, h}[i] \in \left[\frac{k}{2^b}, \frac{k + 1}{2^b}\right[ \right), \\
k &\in \{0,1,...,2^b - 1\} \nonumber
\end{align}

$M$ being the number of images in our dataset that we use to estimate the distribution, here we use 5\% of ImageNet-1K. Thus, we can compute the entropy of every attention weight and quantize their uncertainty as follows:

\begin{align}
\mathcal{H}_{l,h}[i] = -\sum_{k = 0}^{2^b - 1} p^{i}_{l,h}(k)\log_2 (p^{i}_{l,h}(k))
\end{align}

where $\mathcal{H} \in \mathbb{R}^{L\times H \times (N+1) \times (N+1)}$. Through this process, we derive a matrix that encapsulates the uncertainty and redundancy for each weight in the attention maps.

\subsubsection{Visualization}

First, we visualize the entropy maps of the class token ($CLS$), specifically focusing on the first row of the entropy map, for the first six attention heads of the DeiT-Tiny model, as shown in Figure \ref{fig:viz_eam}.

These visualizations reveal distinct behavioral differences between attention heads, indicating that some heads are more redundant than others. For instance, layer 2 head 1 (L2H1) exhibits entropy values near zero for most of its elements, thereby confirming our initial hypothesis. Importantly, this low entropy does not imply that the head is less important or can be discarded, rather, it suggests that its values are highly certain and stable. In fact, some heads, such as L2H1, demonstrate low entropy, indicating that their weights remain stable.

Examining a full entropic attention map in Figure \ref{fig:viz_full_eam} provides a clearer view of the uncertainty within an attention head. Once again, we observe distinct behaviors between two different heads. In particular, the matrix for L2H3 appears almost entirely empty of information, indicating high redundancy and predictability in many of its attention weights.

\begin{figure*}[!t]
    \centering
    \begin{subfigure}[b]{0.30\textwidth}
        \centering
        \includegraphics[width=\textwidth]{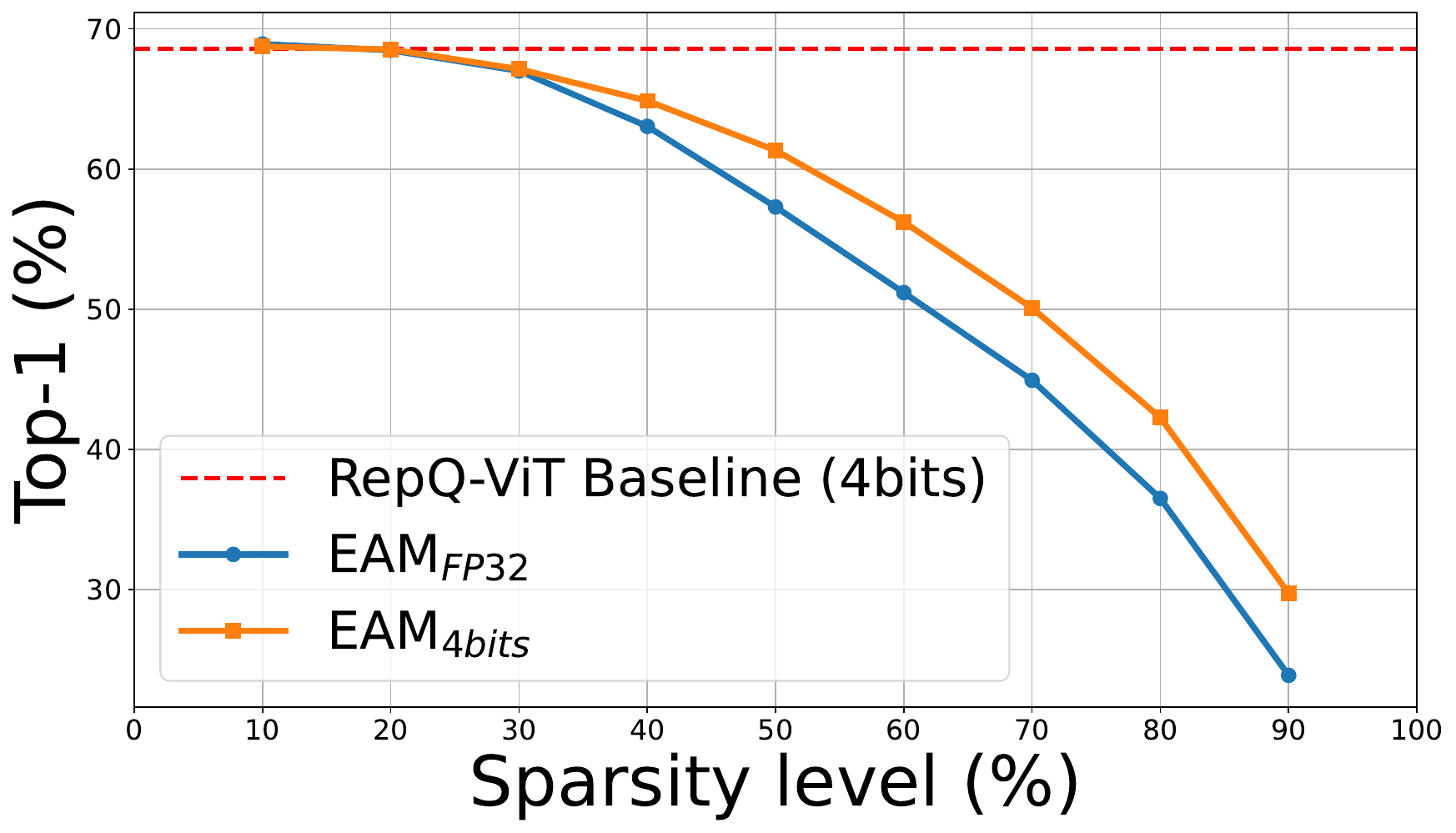}
        \caption{DeiT-Small}
        \label{fig:image2}
    \end{subfigure}
    \hfill
    \begin{subfigure}[b]{0.30\textwidth}
        \centering
        \includegraphics[width=\textwidth]{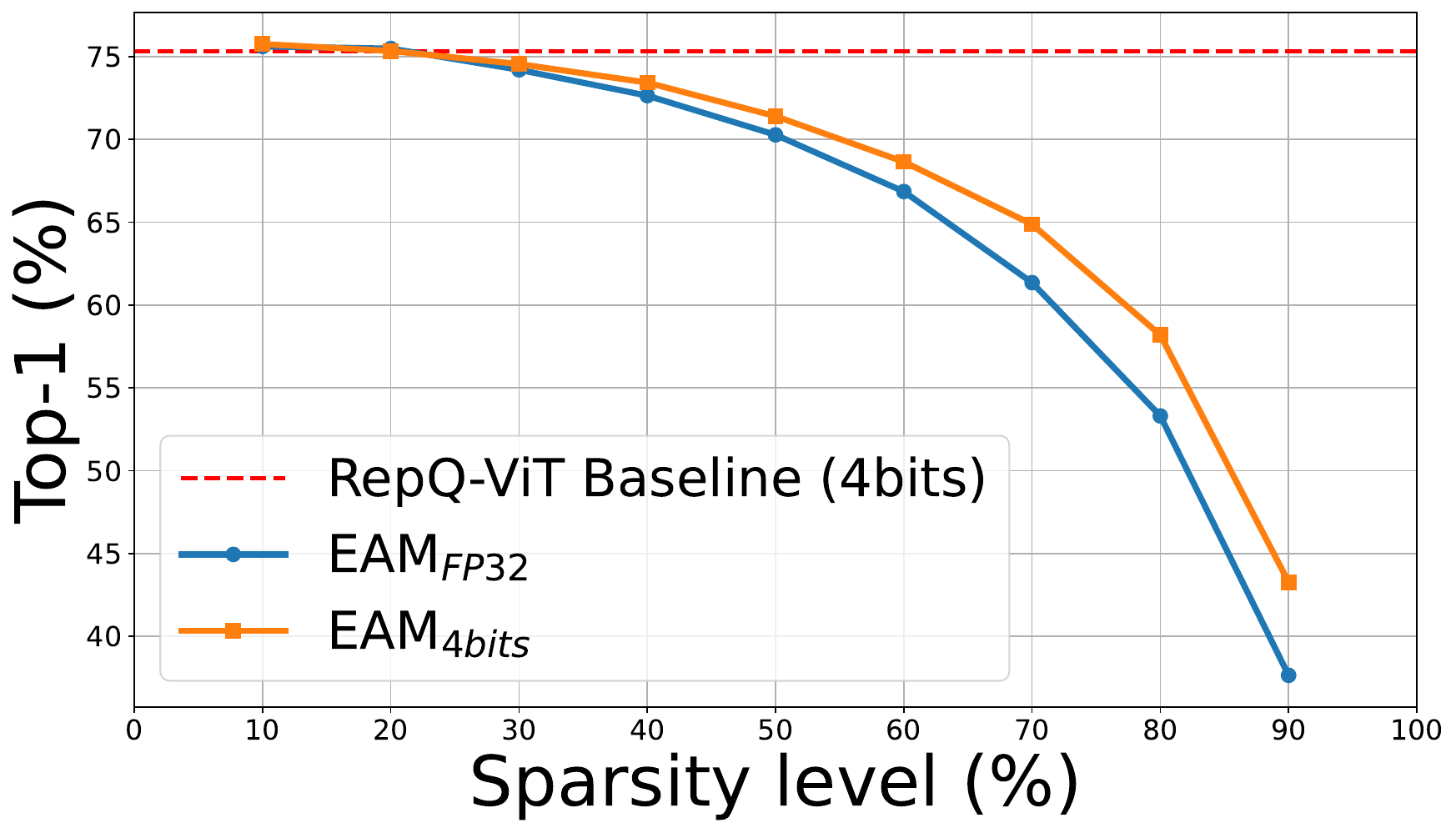}
        \caption{DeiT-Base}
        \label{fig:image3}
    \end{subfigure}
    \hfill
    \begin{subfigure}[b]{0.30\textwidth}
        \centering
        \includegraphics[width=\textwidth]{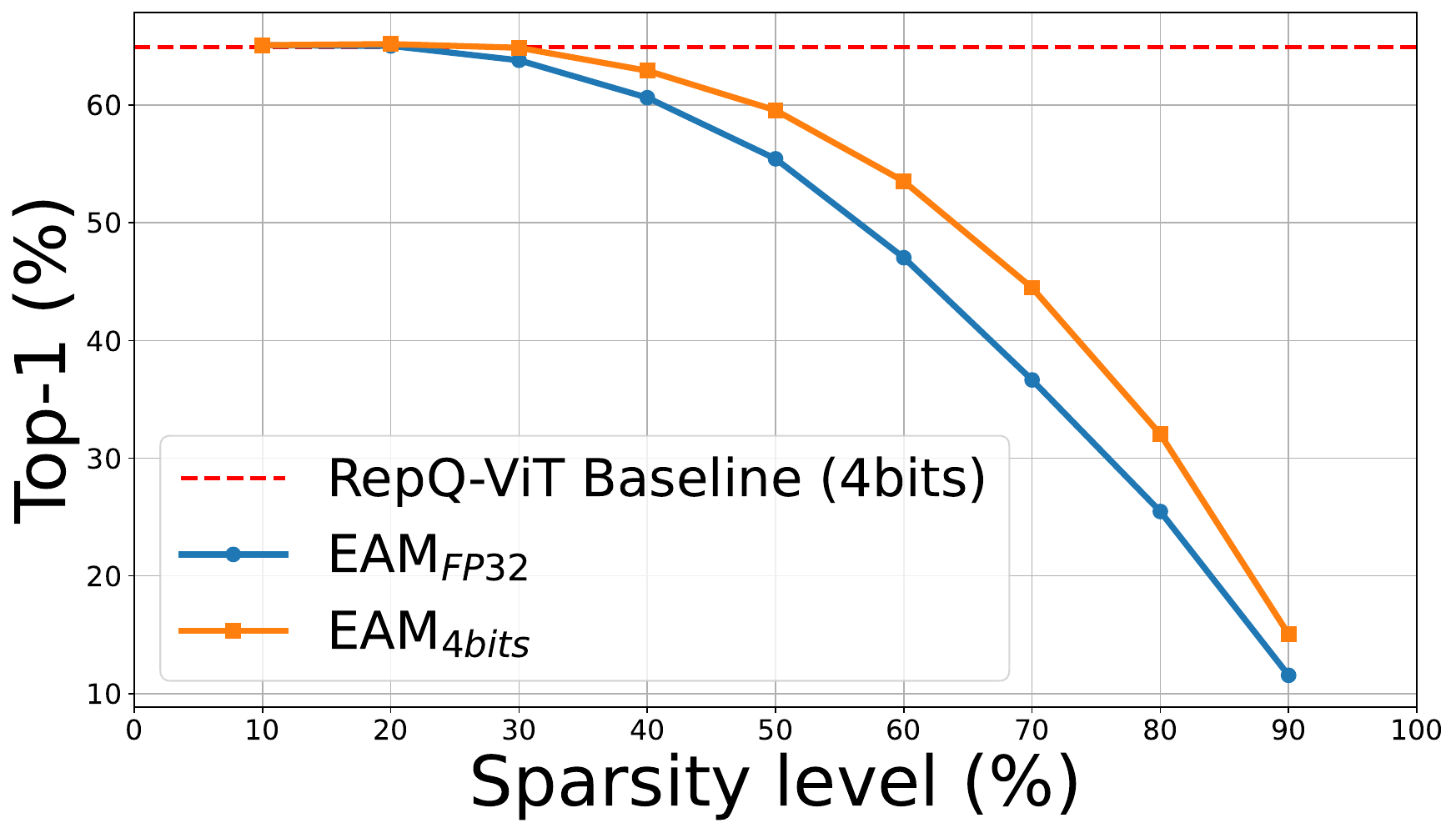}
        \caption{ViT-Small}
        \label{fig:image4}
    \end{subfigure}
    
    \begin{subfigure}[b]{0.30\textwidth}
        \centering
        \includegraphics[width=\textwidth]{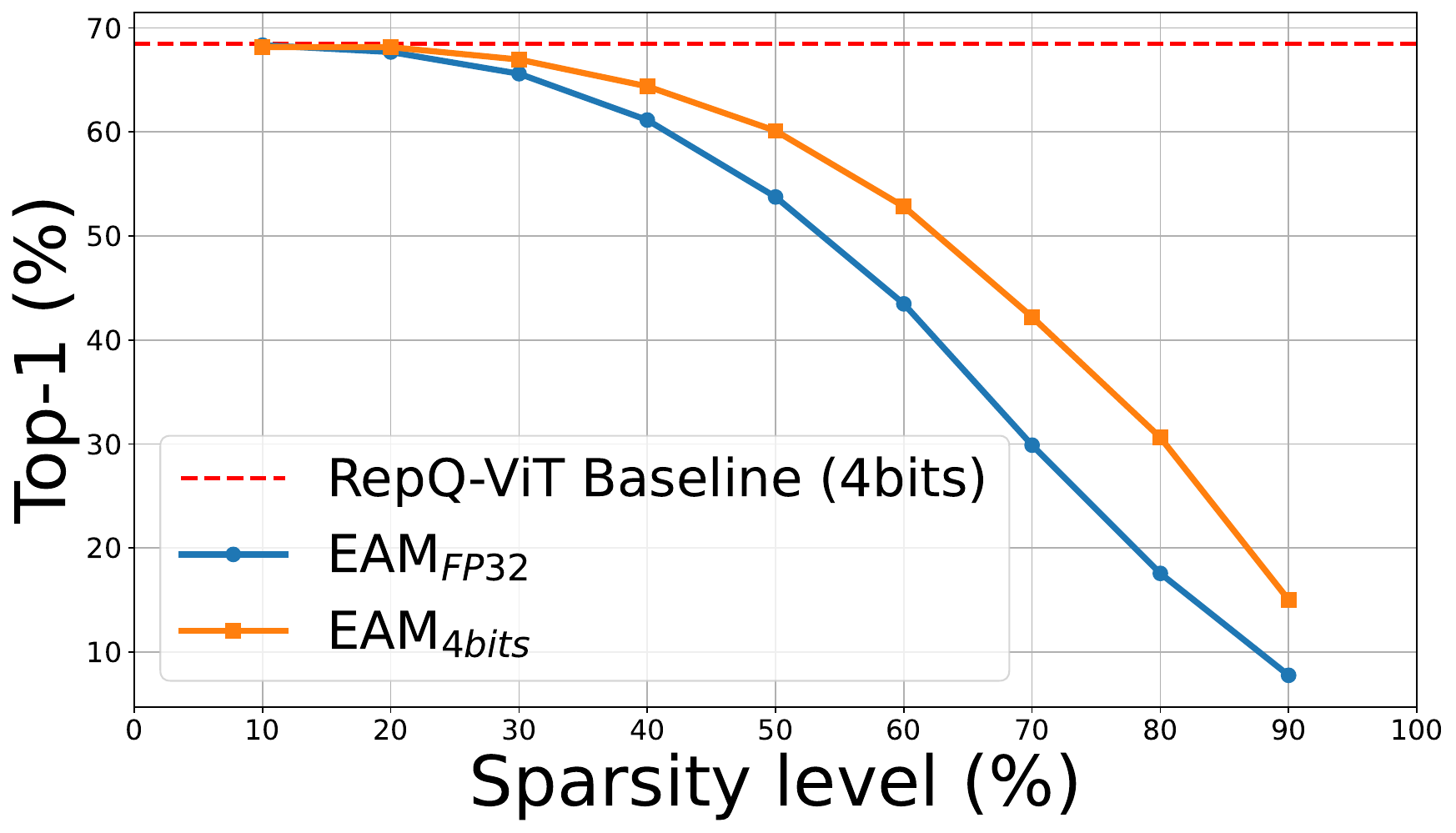}
        \caption{ViT-Base}
        \label{fig:image5}
    \end{subfigure}
    \hfill
    \begin{subfigure}[b]{0.30\textwidth}
        \centering
        \includegraphics[width=\textwidth]{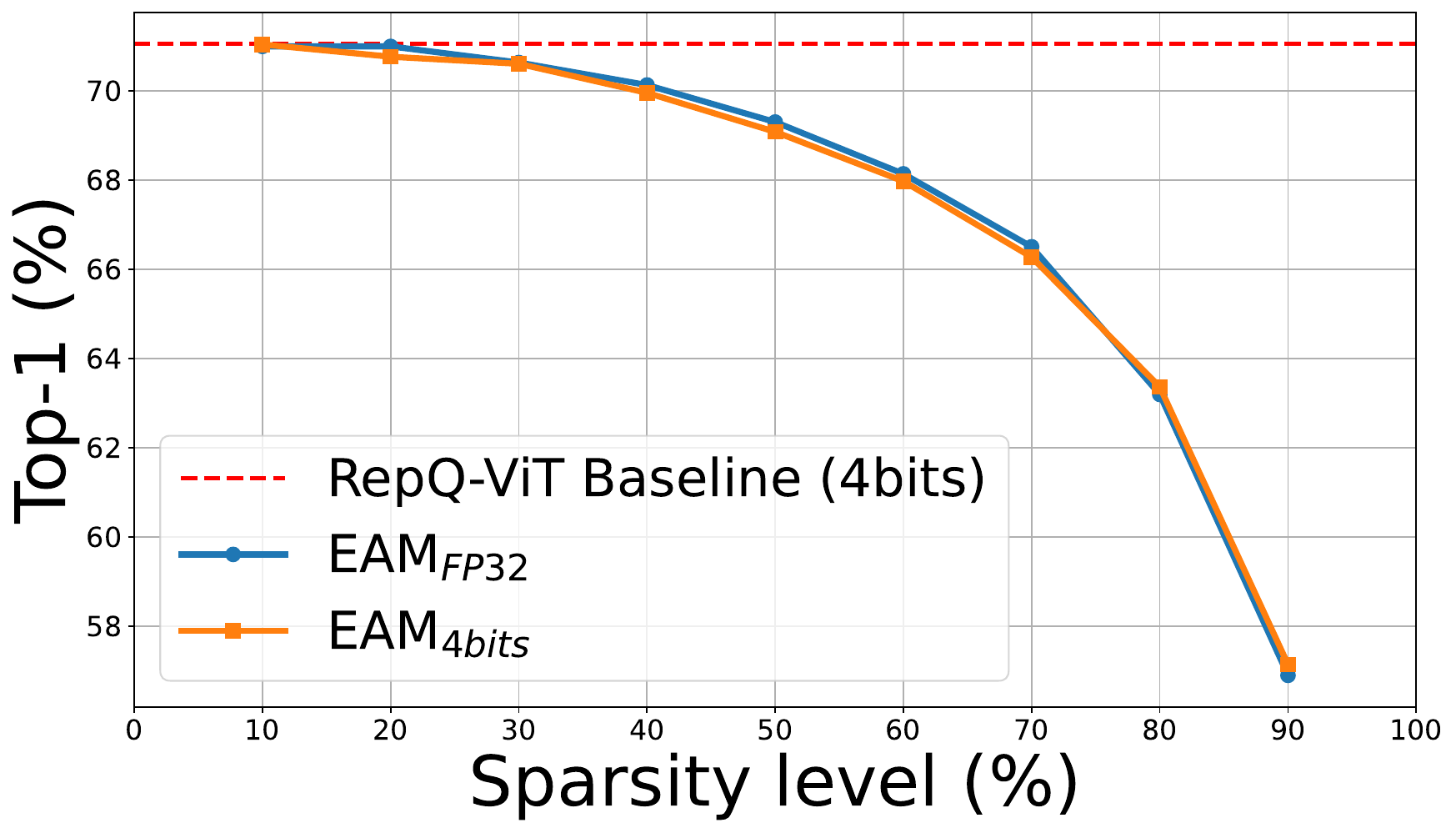}
        \caption{Swin-T}
        \label{fig:image5}
    \end{subfigure}
    \hfill
    \begin{subfigure}[b]{0.30\textwidth}
        \centering
        \includegraphics[width=\textwidth]{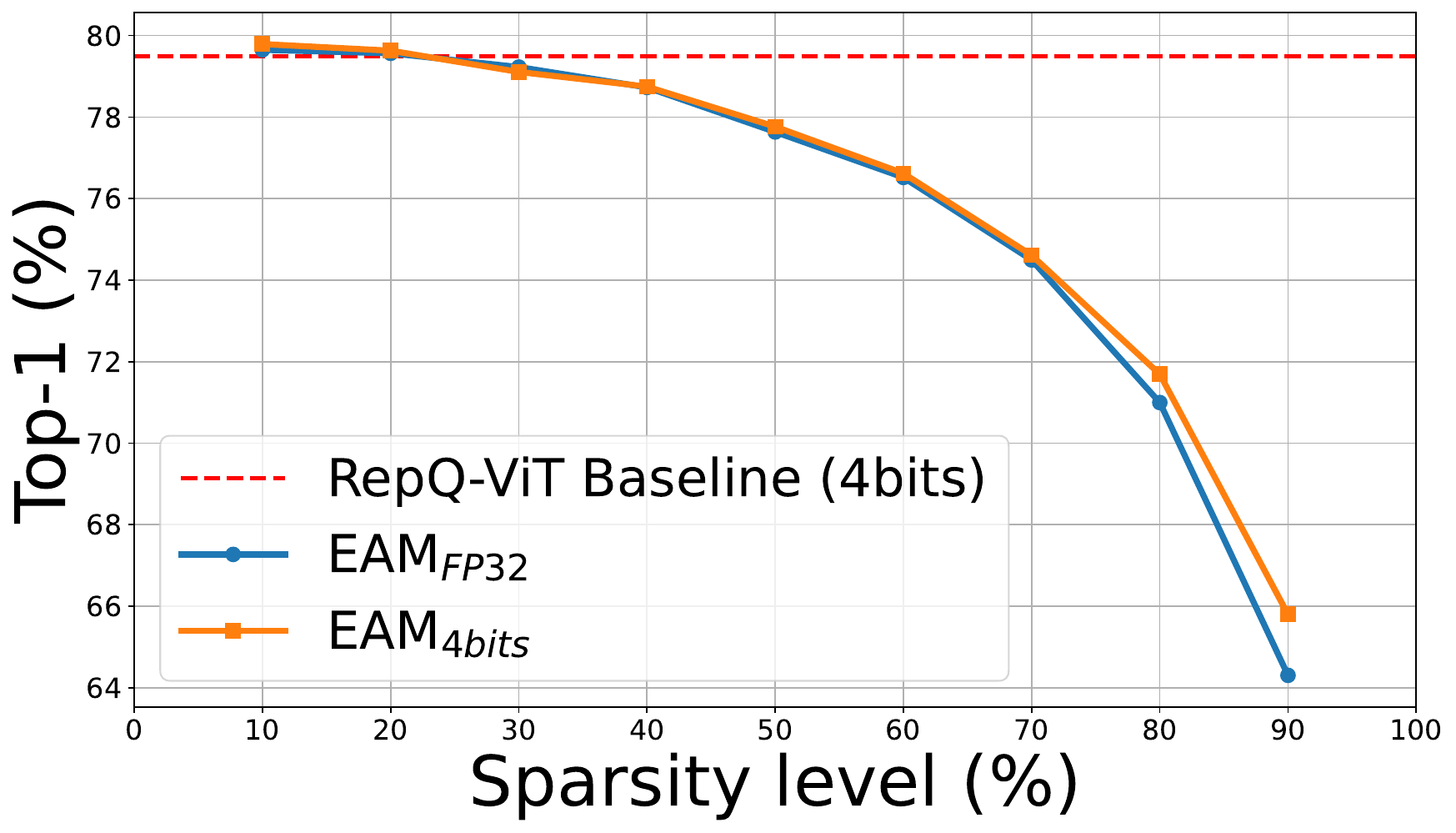}
        \caption{Swin-S}
        \label{fig:image4}
    \end{subfigure}
        \caption{Top-1 accuracy of EAM compared to the RepQ-ViT baseline for  different sparsity levels across various ViT models.}
    \label{fig:perf_vs_sparsity}
\end{figure*}

\subsection{Impact of Quantization}

EAM can also help us understand the impact of quantization on our model. If we see quantization as adding noise to the weights and activations, its impact should be detected in our entropy attention maps.

To better quantify the amount of noise introduced, we can use the KL divergence to measure the distance between two distributions:

\begin{align}
D_{KL}(p_{\text{fp32}}||p_q) = \sum_{x\in \mathcal{X}} p_{\text{fp32}}(x)\log \frac{p_{\text{fp32}}(x)}{p_q(x)}
\end{align}

This formula allows us to compare the distribution of attention weights from the non-quantized model $p_{\text{fp32}}(x)$ with its quantized counterpart $p_q(x)$.

Figure \ref{fig:viz_div} illustrates two examples of divergence maps on the $CLS$ token for a DeiT-Tiny model with 4-bit quantized weights and activations, along with the evolution of the mean and maximum divergence within the model. We observe distinct behaviors among different heads; for instance, the fifth head (L2H3) exhibits higher divergence values compared to the first head. Additionally, quantization appears to have a more significant impact on the initial layers, gradually diminishing towards the end. This suggests that quantization affects each head uniquely, leading us to believe that EAM should be computed on the quantized version to more accurately represent the dynamics of the uncertainty introduced by quantization.

\subsection{Attention weights fixing}

As noted in previous sections, many attention weights are redundant and remain unchanged across different inputs. We can therefore fix these weights and replace them with their mean value. This approach allows us to reduce the computational cost of the model by bypassing a portion of the attention mechanism. A mask is applied to the attention map, where the values are defined as the $\tau\%$ lowest of the entropic attention map:

\begin{align}
A_{l,h}^\text{fix} &= A_{l,h} \otimes (H_{l,h} > \epsilon_\tau) + A^{\mu}_{l,h} \otimes (H_{l,h} < \epsilon_\tau) \label{eq:fixed}, \\
&\mathrel{\phantom{=}} A^{\mu}_{l,h} = \frac{1}{M} \sum_m^M A^{m}_{l,h} \nonumber
\end{align}

%%%MANQUE DETAILLE DES TERMES MATHEMATIQUES%%%%%

In the following sections, we will refer to $\tau$ as the sparsity level, although in the literature this typically denotes matrices with $\tau\%$ zeros. In our context, the matrix $A_{l,h} \otimes (H_{l,h} > \epsilon_\tau)$ becomes sparse, containing zeros that will be replaced by the mean $A_{l,h}^{\mu}$.

\section{Experiments}
\label{sec:Experiments}

\subsection{Set up}

We conduct experiments on common ViT-based models, including ViT-Small, ViT-Base, DeiT-Tiny, DeiT-Small, DeiT-Base, Swin-Tiny, and Swin-Small using the timm package. For post-training quantization, we employ RepQ-ViT \cite{li2023repq}, which demonstrates excellent performance on image classification. We adhere to the original paper's configuration for quantization parameters and use ImageNet-1K to calibrate the entropy of attention maps.

Besides, we investigate the impact of attention sparsity on the accuracy of EAM across these ViT models, and we compare two variants of our model to the RepQ-ViT baseline. EAM$_{\text{FP32}}$, where model weights and activations are quantized to 4 bits while frozen attention map weights retain 32-bit precision, and EAM$_{\text{4bits}}$, where model weights, activations and frozen attention map weights are quantized to 4 bits. Finally, we conduct an ablation study comparing EAM with random fixing, where we use random selection and fixing of attention map weights instead of EAM.

\subsection{Results}

\subsubsection{Impact of sparsity}

Figure \ref{fig:perf_vs_sparsity} illustrates the Top-1(\%) accuracy of EAM on ImageNet-1K across varying sparsity levels for models quantized to 4-bit weights and activations. Two variants of EAM are compared: (1) EAM$_{FP32}$, derived from the full-precision (unquantized) model, and (2) EAM$_{4bits}$, computed using the quantized model via RepQ-ViT.

First, we observe that a sparsity level of up to 30\% can be achieved across all models without significant accuracy degradation. By selectively targeting low-entropy attention weights, these can be frozen without compromising model performance. Second, the EAM$_{4bits}$ computation reduces the performance gap relative to the baseline, particularly at higher sparsity ratios (e.g., $>$ 50\%) compared to EAM$_{FP32}$, this is true for all DeiT models.

Furthermore, at lower sparsity levels (10–20\%), EAM$_{4bits}$ occasionally enhances accuracy compared to the RepQ-ViT baseline. For example, DeiT-Base achieves a Top-1 accuracy of 75.31\% in the baseline configuration but improves to 75.71\% at 10\% sparsity and 75.64\% at 20\% sparsity. At intermediate sparsity levels (30-40\%), the Top-1 accuracy drop is not significant compared to the gains in complexity.

Finally, we observe that Swin-based models behave somewhat differently and appear to be more robust to fixed attention weights. Specifically, they can tolerate up to 50\% sparsity with less than a 2\% drop in accuracy, compared to a 10\% drop for DeiT-Small.

\begin{table*}[!t]
\centering
\renewcommand{\arraystretch}{1.2}
\setlength{\tabcolsep}{4pt}
\begin{tabular}{lcccccccc}
\toprule
\textbf{Method} & \textbf{Prec. (W/A)} & \textbf{ViT-S} & \textbf{ViT-B} & \textbf{DeiT-T} & \textbf{DeiT-S} & \textbf{DeiT-B} & \textbf{Swin-T} & \textbf{Swin-S} \\
\midrule
Full-Precision  & 32/32 & 81.39 & 84.54 & 72.21 & 79.85 & 81.80 & 81.30 & 83.23 \\
RepQ-ViT & 4/4   & 64.92 & \textbf{68.46} & 57.91 & 68.58 & 75.31 & \textbf{70.67} & 79.45 \\
\textbf{EAM${_{4bits}^{\tau=10\%}}$}      & 4/4   & \textbf{65.09} & 68.18 & \textbf{58.03} & \textbf{68.74} & \textbf{75.71} & 70.65 & \textbf{79.79} \\
\textbf{EAM${_{4bits}^{\tau=20\%}}$}      & 4/4   & \textbf{65.19} & 68.16 & \textbf{58.02} & 68.53 & \textbf{75.64} & 70.55 & \textbf{79.63} \\
\bottomrule
\end{tabular}
\caption{Top-1 accuracy of EAM under 4-bit precision on various ViT models for 10\% and 20\% sparsity levels, compared to RepQ-ViT and Full-Precision baselines.}
\label{tab:eah_results}
\end{table*}

\begin{figure}[h]
    \centering
    \begin{minipage}{0.23\textwidth}
        \centering
        \includegraphics[width=\linewidth]{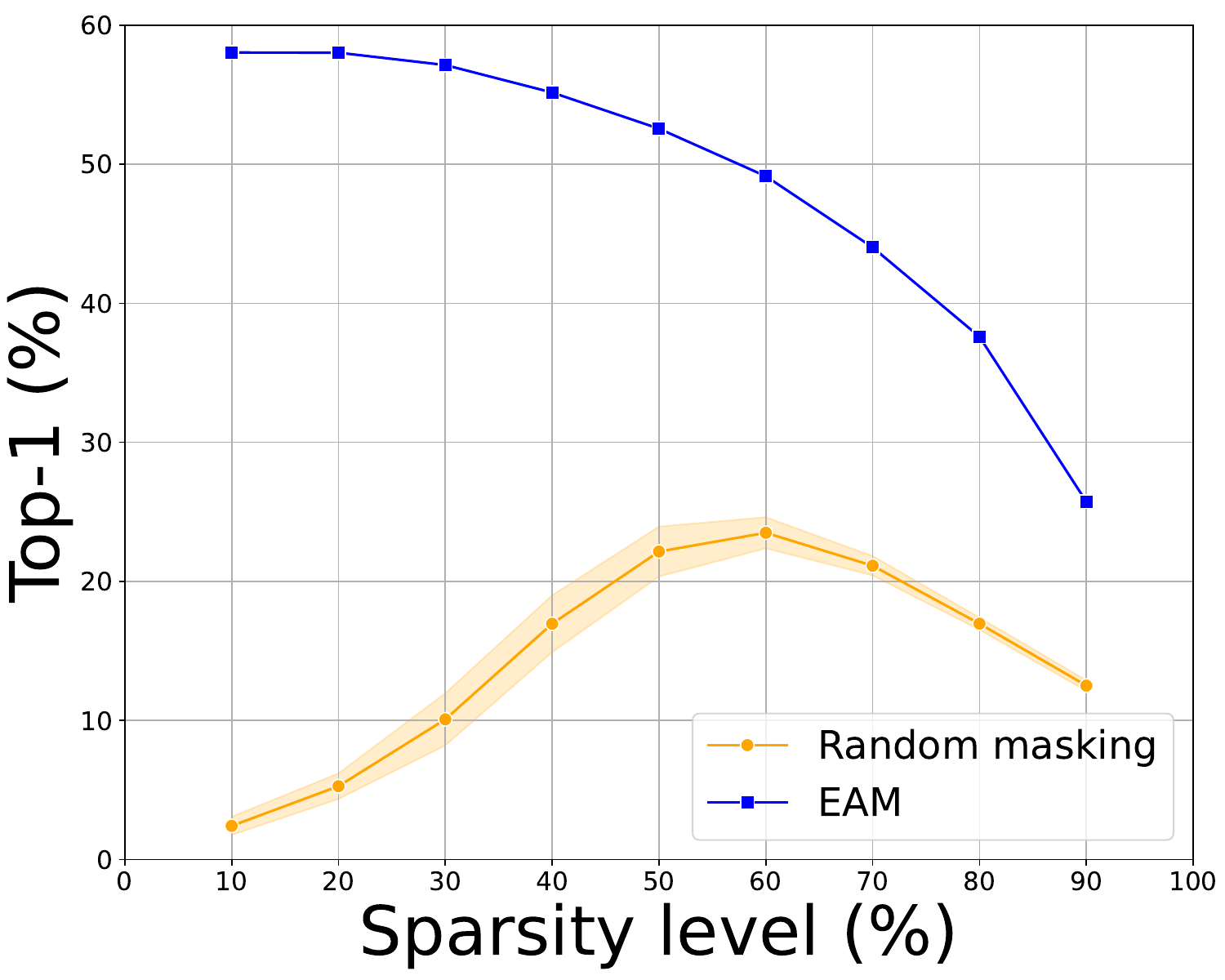}
        \subcaption{}
        \label{fig:a}
    \end{minipage}
    \hfill
    \begin{minipage}{0.23\textwidth}
        \centering
        \includegraphics[width=\linewidth]{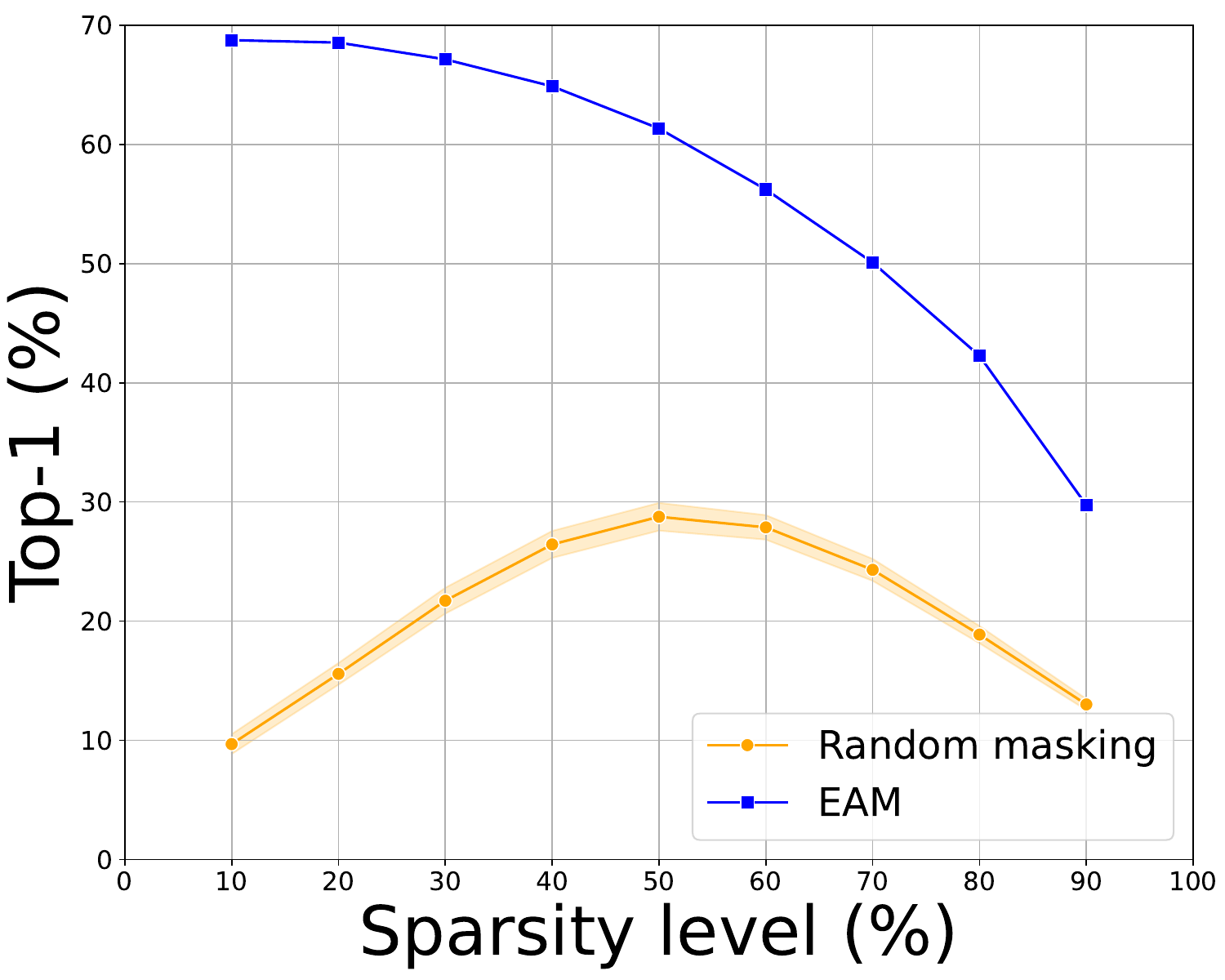}
        \subcaption{}
        \label{fig:b}
    \end{minipage}
    
    \begin{minipage}{0.23\textwidth}
        \centering
        \includegraphics[width=\linewidth]{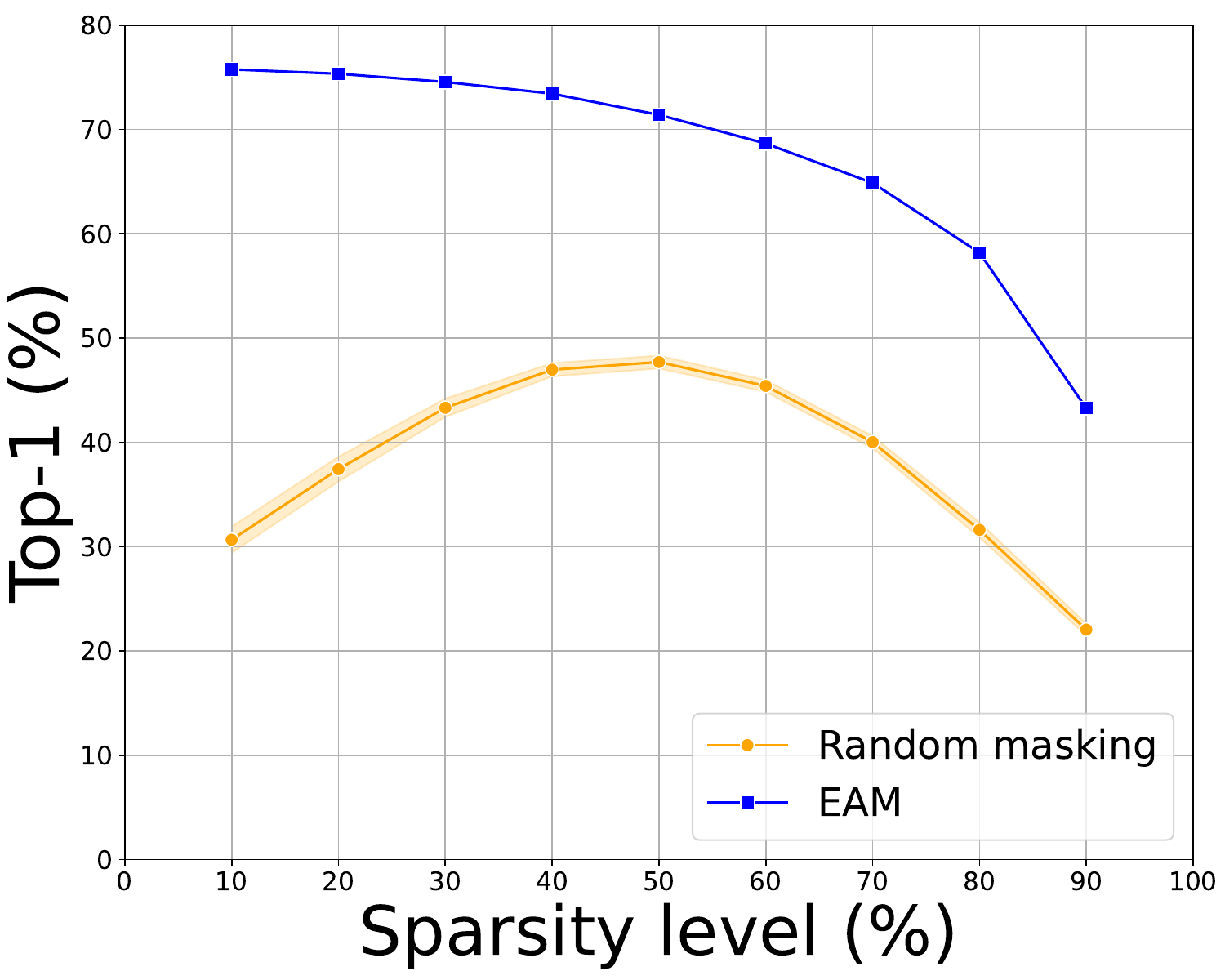}
        \subcaption{}
        \label{fig:c}
    \end{minipage}
    \hfill
    \begin{minipage}{0.23\textwidth}
        \centering
        \includegraphics[width=\linewidth]{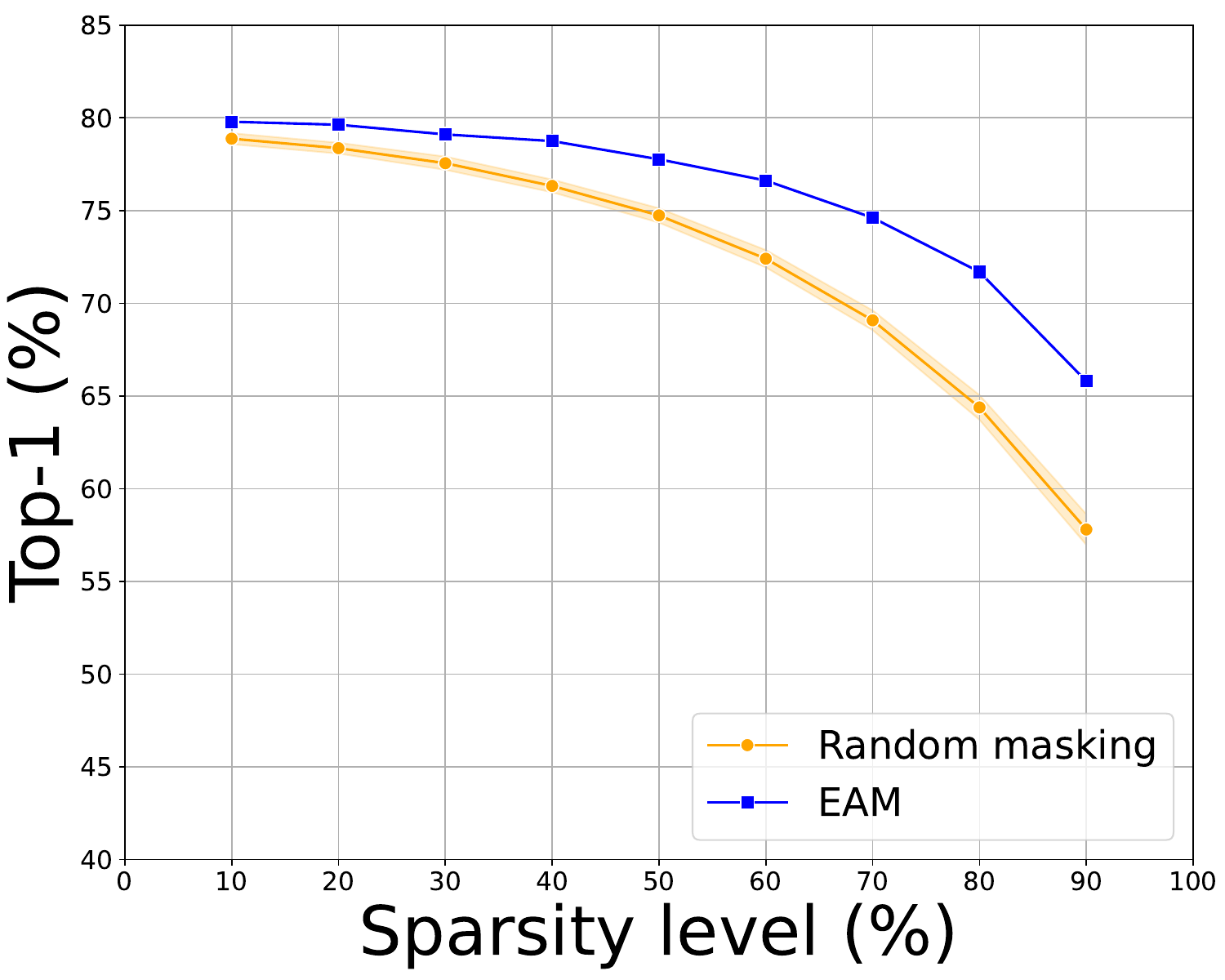}
        \subcaption{}
        \label{fig:c}
    \end{minipage}
    \caption{Ablation study of EAM against random fixing across sparsity levels. (a) DeiT-Tiny, (b) DeiT-Small, (c) DeiT-Base, (d) Swin-S.}
    \label{fig:ablation_study}
\end{figure}

Table \ref{tab:eah_results} shows the performance of EAM with 4-bit precision at $\tau=10\%$ and $\tau=20\%$ sparsity levels. In general, EAM leads to an increase in performance at these sparsity levels across all models except ViT-Base and Swin-Tiny, where the accuracy drop is minimal, at less than 0.3\% and 0.12\%, respectively. Although the improvement is modest, it is worth noting that fixing and quantizing the attention weights to 4 bits can contribute to enhanced performance. The following section will compare our entropy-based fixing of attention weights in EAM with random fixing.

\subsubsection{Random fixing}

%%% Version du L.
% To validate the effectiveness of our masking approach, we compare it against randomly generated masks. Fixing the sparsity ratio at 10\%, we permute the initial EAM-computed mask across 10 different random seeds. As shown in Table \ref{tab:random}, which presents the Top-1 accuracy comparison between our EAM-derived masks and random masks, our method consistently outperforms random masking across all evaluated models. The empirical results demonstrate that structured attention-based masking provides superior performance compared to unstructured random sparsity patterns.

%%Proposition du K. Au lieu de fixer le ratio à 10%, on met une figure sur tous les ratio [0.1, 0.9] pour EAM et random, et ce sur les trois modèles de deit. ça a plus de sens, c'est plus visuel, et ça soutient EAM. Aussi, un petite analyse des résultats, surtout quand on voit que le random n'a pas le même effet sur tiny que sur base (et on sait pourquoi)

To validate the effectiveness of our entropy-based fixing approach in EAM, we compare it against random fixing, i.e., randomly selected attention weights. Figure \ref{fig:ablation_study} illustrates the Top-1 accuracy versus the sparsity level of EAM against random fixing across DeiT-Tiny, DeiT-Small, DeiT-Base, and Swin-S, where our method consistently outperforms random fixing on all these models. 

The main insights of this ablation study are two-fold. First, the discrepancy between the accuracy of EAM and random fixing is significant enough to validate our approach, especially on DeiT models, where Top-1 accuracy of random fixing collapses, keeping the gap to our model as low as 11.67\% and as high as 55.60\% on DeiT-Tiny. Although the gap is less pronounced on Swin-S, it remains statistically significant across sparsity levels, as the gap in Top-1 accuracy increases from 0.91\% at 10\% sparsity to 6.60\% at 90\% sparsity.

Second, DeiT is more sensitive to random fixing than Swin. This difference likely arises from the localized attention windows and hierarchical structure in Swin: While DeiT relies on global attention across all patches, Swin restricts interactions to local regions and refines features through downsampling. As a result, randomly fixing parts of attention maps in DeiT shows a higher impact on the subsequent layers compared to Swin, as the damage is confined to a single window, and later layers can recover lost information through merged features.

\section{Conclusion}
\label{sec:Conclusion}

In this work, we introduced EAM, an entropy-driven approach to optimize Vision Transformers by analyzing and exploiting the information redundancy in attention heads. Our main insight is that low-entropy attention heads exhibit stable, predictable patterns across inputs, allowing us to fix and quantize them aggressively without compromising model performance. Through extensive experiments on ImageNet-1K with various ViT architectures, we demonstrated that our method reduces computational complexity and memory demands while maintaining accuracy. Specifically, EAM increases the Top-1 accuracy of the RepQ-ViT baseline while fixing 10\% to 20\% of the weights in attention maps and quantizing them to 4 bits. Furthermore, we achieved up to 40\% sparsity in attention maps with negligible performance degradation. Finally, we validated the entropy-based fixing in EAM with an ablation study with random fixing, and showed that EAM outperforms random fixing on all the ViT models. In future work, we will extend our method to larger architectures, including Vision-Language Models (VLMs) and Large Language Models (LLMs), which process longer context sequences compared to ViTs. Given the high computational cost of MHSA, applying our method to these models is expected to yield greater computational savings, therefore, we find it interesting to validate these gains experimentally. In addition, we can further extend our work by enabling the model to retrain with attention weights fixed with EAM, aiming to minimize the loss in accuracy.

% % Perspectives
% \begin{itemize}
%     \item Scale to high computational models such as LLMs and VLM (long context windows)
%     \item Experiment on SSMs and Mamba?
%     \item BumbleBee ?
% \end{itemize}

% \input{sec/2_formatting}
% \input{sec/3_finalcopy}
{
    \small
    \bibliographystyle{ieeenat_fullname}
    \bibliography{main}
}

\end{document}